\documentclass[10pt,twocolumn,letterpaper]{article}
\usepackage[final]{wacv} 

\usepackage{graphicx}
\usepackage{amsmath}
\usepackage{amssymb}
\usepackage{booktabs}

\usepackage{multirow}
\usepackage{colortbl}

\usepackage[pagebackref,breaklinks,colorlinks,bookmarks=false]{hyperref}
\usepackage[accsupp]{axessibility}  

\usepackage[capitalize]{cleveref}
\crefname{section}{Sec.}{Secs.}
\Crefname{section}{Section}{Sections}
\Crefname{table}{Table}{Tables}
\crefname{table}{Tab.}{Tabs.}

\def\wacvPaperID{2265} 
\def\confName{WACV}
\def\confYear{2024}

\begin{document}

\title {Text-Guided Face Recognition using Multi-Granularity Cross-Modal Contrastive Learning}  

\author{Md Mahedi Hasan, Shoaib Meraj Sami, and Nasser Nasrabadi\\
	West Virginia University, Morgantown, West Virginia, USA\\
	{\tt\small mh00062@mix.wvu.edu, sms00052@mix.wvu.edu, nasser.nasrabadi@mail.wvu.edu}
}

\maketitle

\begin{abstract}
State-of-the-art face recognition (FR) models often experience a significant performance drop when dealing with facial images in surveillance scenarios where images are in low quality and often corrupted with noise. Leveraging facial characteristics, such as freckles, scars, gender, and ethnicity, becomes highly beneficial in improving FR performance in such scenarios. In this paper, we introduce text-guided face recognition (TGFR) to analyze the impact of integrating facial attributes in the form of natural language descriptions. We hypothesize that adding semantic information into the loop can significantly improve the image understanding capability of an FR algorithm compared to other soft biometrics. However, learning a discriminative joint embedding within the multimodal space poses a considerable challenge due to the semantic gap in the unaligned image-text representations, along with the complexities arising from ambiguous and incoherent textual descriptions of the face. To address these challenges, we introduce a face-caption alignment module (FCAM), which incorporates cross-modal contrastive losses across multiple granularities to maximize the mutual information between local and global features of the face-caption pair. Within FCAM, we refine both facial and textual features for learning aligned and discriminative features. We also design a face-caption fusion module (FCFM) that applies fine-grained interactions and coarse-grained associations among cross-modal features. Through extensive experiments conducted on three face-caption datasets, proposed TGFR demonstrates remarkable improvements, particularly on low-quality images, over existing FR models and outperforms other related methods and benchmarks.
\end{abstract}


\section{Introduction}
\label{sec:intro}
Current one-to-one face recognition (FR) algorithms such as ArcFace~\cite{arcface} and AdaFace~\cite{adaface} face challenges in achieving high verification rates (VR), particularly in surveillance-based applications~\cite{H_2017, Yin_2020}, where a low-resolution face probe is captured under non-ideal conditions. Covariate factors like non-uniform lighting, occlusion, and non-frontal pose in such scenarios, severely degrade the image quality, leading to decreased matching performance~\cite{Li_2019_low, Yin_2020}. One way to improve the performance of an FR algorithm is to integrate auxiliary information, such as facial marks, gender, age, skin and hair color, facial expression, and other distinctive facial attributes to a face recognition model~\cite{Jain_2009, Gonzalez_2018, Zhang_2015}.

Natural language descriptions, which highlight the distinct characteristics of a face image, can also be employed as a soft biometric to improve the performance of a FR model~\cite{cgfr}. In this research, we hypothesize that semantic information can substantially improve the image understanding capability of an FR algorithm compared to other soft biometrics. Therefore, the core idea of this research is to develop a model that effectively integrates textual descriptions with facial images, thereby maximizing performance. As both image and textual modalities are complementary to each other, the combination of textual and facial features will result in a significant performance leap. This text-guided face recognition (TGFR) has great potential for applications in systems where face images are captured in real-world settings, such as criminal investigations and intelligent video surveillance~\cite{cgfr}. In criminal investigations, the testimony of the witnesses can be effectively utilized within the TGFR model to find or narrow down the potential suspects even from low-quality images.


However, integrating textual descriptions or captions within a FR loop poses several challenges. Firstly, natural language captions are inherently more abstract than facial images. A limited set of facial attributes may be insufficient to convey all the fine-grained details of a complex face image. Secondly, as describing a face image involves subjective judgment, captions written by different annotators for a particular face can be completely or partially different. Sometimes, an attribute, which one annotator considers important may be ignored by others. Thirdly, different captions of a particular face image may contain large word variability. Fourthly, annotators might disagree on specific attributes, leading to incorrect representations of a person's face. For example, annotators may differ on whether the person has a sharp nose. Lastly, annotators occasionally include a surplus of non-discriminative information, readily present on the face, thus making it challenging to differentiate the face-caption pair from others. Therefore, learning accurate embedding from such textual descriptions is the most challenging step of a text-guided FR model.


As the primary objective of this work is to improve the performance of SOTA FR algorithms~\cite{arcface, adaface, magface}, in our TGFR framework, we utilize them as fixed feature extractor. By keeping them in a frozen state, we can accurately measure the performance gain of the FR model resulting from textual supervision. However, the contextual embeddings generated from the foundation model are unaligned with image features and have limited distinctiveness for face recognition. To address these challenges, we propose a face caption alignment module (FCAM) that finetunes the text encoder with the specific aim of learning text embeddings that are both discriminative and well-aligned with visual features. The proposed FCAM  incorporates a global caption-image contrastive loss (CICL) and a local word-region contrastive loss (WRCL) to learn alignments between the caption-image and word-regions, respectively. Additionally, it includes an intra-modal contrastive objective as well as an identity loss to produce discriminative features for both visual and semantic modalities. 


Furthermore, in a conventional dual-encoder architecture, a fusion scheme is typically employed to learn a joint representation from image and text features. However, a simple feature-level fusion (FLF) scheme may be ineffective as the regional image features and word embeddings lack prior relationships~\cite{ALBEF}. Moreover, the FLF scheme treats textual and image features as separate entities, thus failing to capture their interactions and dependencies. Therefore, we introduce a novel and efficient attention-guided fusion scheme called face-caption fusion module (FCFM), aiming to explore cross-modal fine-grained interactions and coarse-grained associations for improved performance in multimodal fusion. In summary, our contributions are three-fold:


\begin{itemize}
	\item We boost the performance of SOTA FR algorithms, including AdaFace~\cite{adaface} and ArcFace~\cite{arcface}, by learning effective representations through cross-modal and intra-modal contrastive supervision.
	\item We introduce FCAM, a novel module to learn aligned global and local features within a shared semantic space by effectively utilizing the CICL to align caption-image embeddings and WRCL to align attention-weighted region with word embeddings.
	\item Experiments conducted on three challenging face-caption datasets demonstrate significant improvements over SOTA methods and baseline approaches in both 1:1 verification and Rank-1 identification.
\end{itemize}

\begin{figure}[t]
	\centering
	\includegraphics[width=0.44\textwidth]{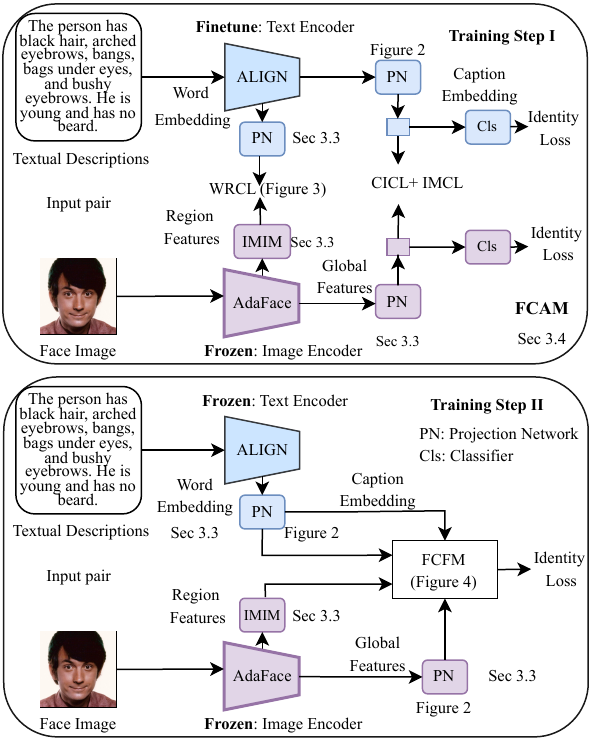}
	\caption{An overview of our proposed TGFR framework, comprising an image encoder, a text encoder, a face-caption alignment module (FCAM), and a face-caption fusion module (FCFM). We follow a two-step learning process. In the initial step, we train both the foundation model and the FCAM module while keeping the parameters of the FR model frozen. The goal of FCAM is to map the local and global embeddings of face-caption pairs into a shared space, ensuring a proper alignment between image-text modalities. In the final step, we train the FCFM module.}
	\label{fig:overview}
	\vspace{-2mm}
\end{figure}

\section{Related work}
\paragraph{Text-guided face recognition}
Facial semantic attributes have been extensively exploited as auxiliary information for various tasks, such as face image retrieval~\cite{Zaeemzadeh_2021}, generation~\cite{Sun_2022}, and editing~\cite{He_2019, MMCelebA}. Gonzalez~\etal~\cite{Gonzalez_2018} presented an overview of soft biometrics for face recognition in unconstrained scenarios. The authors reported a maximum relative performance improvement of 40\% over the FR models when utilizing 6 discriminative facial attributes.

Our work is closely related to a recent approach to face recognition using captions, namely CGFR, introduced by Hasan and Nasser~\cite{cgfr}. In CGFR, the authors employed a pre-trained FR model and a pre-trained BERT model~\cite{bert} for extracting features from the respective modalities. They further finetuned the BERT model using a refinement module. Afterward, a cross-modal fusion scheme was employed to transform the uni-modal representations into a unified embedding space. A notable drawback of CGFR~\cite{cgfr} is that its feature extractors were trained independently on objectives that were distinct and unrelated to the FR task. For example, language models such as BERT~\cite{bert} and RoBERTa~\cite{roberta} were trained on vast corpora using masked language modeling. Therefore, these encoders generate irrelevant features that result in lower performance in downstream FR tasks.

In this research, rather than adopting BERT, we employ a CLIP-like foundation model~\cite{clip, BLIP, align} that has been pre-trained on a large-scale dataset containing image-text pairs for a wide range of multimodal tasks including image-text matching~\cite{uniter, align, ALBEF, fu_2023_learning, wen_2020_learning}, cross-modal retrieval~\cite{ViLBERT, ALBEF}. The rationale behind selecting a CLIP-like model, such as ALIGN~\cite{align}, as our text encoder lies in its primary objective of aligning images and text. Furthermore, they trained CGFR using the SOTA CMPC~\cite{Zhang_2018} and DAMSM~\cite{damsm} losses on the low-quality MMCelebA dataset~\cite{MMCelebA}. In contrast, we train our TGFR using a variety of cross-modal contrastive losses and evaluate our model on three challenging face-caption datasets. Additionally, we incorporate an intra-modal contrastive supervision and an identity loss to further enhance the discriminability of the learned features.

\paragraph{Representation learning with text supervision}
In recent years, with the availability of datasets containing large corpora of image-text pairs, representation learning with text supervision~\cite{align, virtex, ViLBERT, ALBEF, uniter} has exhibited remarkable success across a multitude of downstream tasks such as visual question answering task~\cite{VLBERT, VLMo}, cross-modal retrieval~\cite{ViLBERT, ALBEF}, and and zero-shot image classification~\cite{align}. The semantic context derived from textual supervision aids the model in capturing more comprehensive multimodal representations, resulting in improved performance. Desai and Johnson introduced the VirTex framework~\cite{virtex}, where they trained a CNN model with textual descriptions for image captioning tasks. By fine-tuning the pre-trained model, they achieved improved performance in image classification and instance segmentation tasks. Additionally, in the iCAR framework~\cite{icar}, authors proposed a novel deep fusion approach for effective image-text alignment. This method demonstrated superior performance on several benchmark datasets for zero- and few-shot image classification.

\begin{figure}[t]
	\centering
	\includegraphics[width=0.46\textwidth]{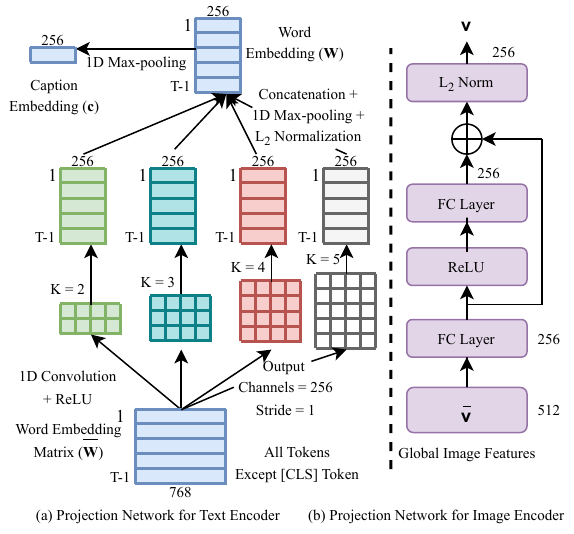}
	\caption{(a) The architecture of the proposed CNN-based projection for generating phrase-level contextual embeddings from the output of the foundation model. (b) The network for projecting global image features into a shared space. (K = 1D kernel)}
	\label{fig:projection_head}
	\vspace{-2mm}
\end{figure}

\paragraph{Contrastive learning} 
To tackle the challenges of computational complexities, noise-contrastive estimation (NCE) was introduced~\cite{nce_loss}. In this method, rather than computing the probability of the target word among all possible words in the vocabulary, a binary cross-entropy loss is employed to differentiate between the actual target word and noise samples.  Another approach, information NCE~\cite{infoNCE}, focuses on maximizing a lower bound on the mutual information existing between images and captions, particularly in the context of cross-modal retrieval tasks. 


Another powerful technique for learning image-text representations is contrastive pretraining, which has recently demonstrated remarkable success in various multimodal tasks~\cite{clip, align, ALBEF}. CLIP~\cite{clip} introduces an approach for learning visual representations using caption by establishing semantic similarity between image-text pairs through contrastive loss. Jia~\etal~\cite{align} designed a method called noisy student training, which utilized noisy text data to improve image-text representation learning by aligning image-text pairs using a contrastive loss. In another approach, authors~\cite{ALBEF} employed the momentum distillation technique for image-text representation learning. They proposed an image-text contrastive loss to align image-text features before fusion. Although these methods achieve remarkable performance on tasks related to multimodal representation learning, they often fail to capture complex interactions between image regions and words, resulting in lower performance on fine-grained image-text classification~\cite{Vilt}.

Moreover, in addition to image-text matching loss~\cite{ViLBERT, clip, align, ALBEF}, numerous other contrastive learning-based cross-modality loss functions have been proposed in the literature. These encompass, triplet loss~\cite{triplet_cons_learning, unimo}, word-region alignment~\cite{damsm, uniter}, word-patch alignment~\cite{Vilt}, token-wise alignment~\cite{mgca} and more, all aimed at tackling a variety of multimodal tasks. For example, Zhang~\etal~\cite{convirt} introduced ConVIRT to learn visual features from chest X-ray images through the analysis of paired reports. They employed a bidirectional contrastive objective between visual and textual modalities, similar to our CICL, and achieved superior performance across four medical image classification and two zero-shot retrieval tasks. Moreover, Huang~\etal~\cite{gloria} contrasted image sub-regions with the corresponding words in the textual report to learn both global and local representations in their proposed GLoRIA framework.

\begin{figure*}[t]
	\centering
	\includegraphics[width=\textwidth]{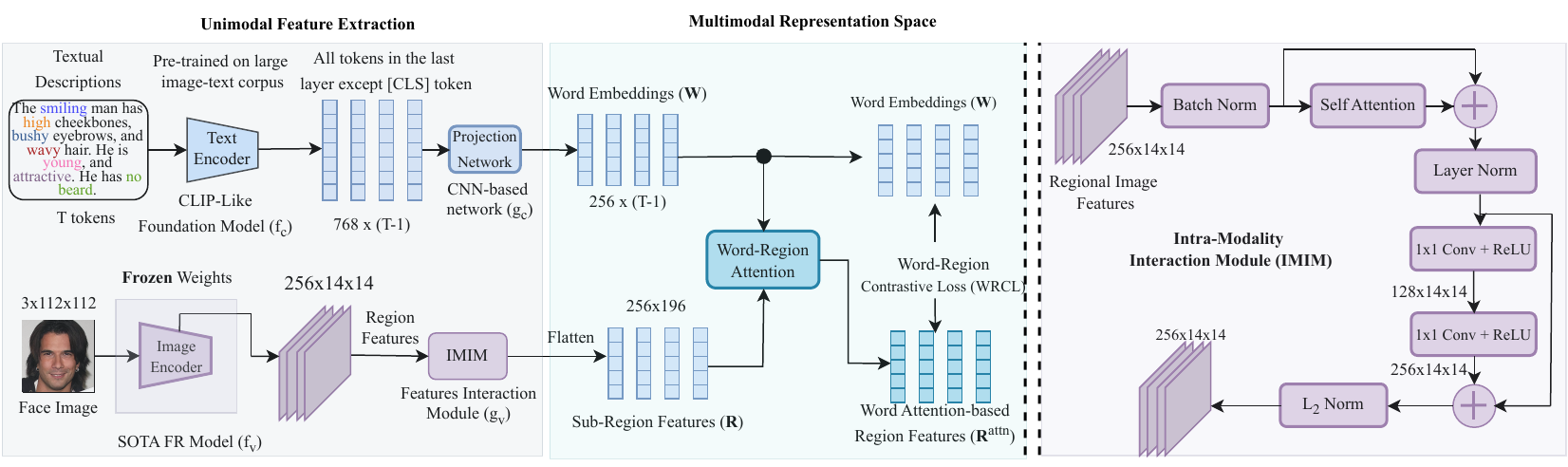}
	\caption{The block diagram of the proposed face-caption alignment module where we learn context-aware sub-regional features by emphasizing the distinctive attributes of a face image, corresponding to a specific word. Initially, the extracted local image features and word embeddings are fed into the projection networks to map them in a shared multimodal semantic space. Next, a word-region contrastive loss is applied to the context-aware regional features and their respective word representations.}
	\label{fig:word_region_cl}
\end{figure*}

\section{Our approach}
\subsection{Model architecture} 
An overview of our method is illustrated in Figure~\ref{fig:overview}. It consists of several components: an image encoder for extracting both local and global image features from an input face, a text encoder for learning embeddings from the associated captions, a face-caption alignment module, and a face-caption fusion module. The uni-modal features are then fed into the FCAM module, which serves three main purposes: 1) aligning cross-modal features, 2) applying intra- and inter-modal feature interactions, and 3) learning discriminative image-text features. Finally, the FCFM module is applied to implement fine-grained interactions between cross-modal features.

\subsection{Multimodal feature extraction}
In this study, we employ a SOTA FR model~\cite{arcface, adaface} as our image encoder, $f_{v}$, and a foundation model~\cite{clip, align, BLIP, flava}, pre-trained on a large number of caption-image pairs, as our text encoder, $f_{c}$. The iRestNet18~\cite{resnet, arcface} is adopted as the backbone network for the image encoder. In this network, the output of the final adaptive average pooling layer produces the global feature vector, $\bar{\textbf{v}} \in \mathbb{R}^{512}$, for the input image which contains high-level semantic information of the face. Additionally, regional image features, $\bar{\textbf{R}} \in \mathbb{R}^{256\times 14\times 14}$, are extracted from an intermediate convolutional stage. From the output of the foundation model, we obtain a contextualized word matrix, $\bar{\textbf{W}} \in \mathbb{R}^{D\times T}$. Here, $T$ represents the total token count, while $D$ denotes the dimensionality of each token embedding.

\subsection{Projection networks (PNs)}
The output of the text encoder comprises 768-dimensional contextualized embeddings for each token in the caption~~\cite{align, BLIP}. In this work, we design a 1D CNN-based projection network to map each embedding into a shared 256-dimensional multimodal space, Figure~\ref{fig:projection_head} (a), illustrates the proposed network which allows us to capture phrase-level contextual information. For convolutions, each utilizing a kernel size ranging from 2 to 5 and a stride of 1 are applied to the word matrix, $\bar{\textbf{W}} \in \mathbb{R}^{768 \times (T-1)}$, to effectively capture local patterns within the text. Next, we concatenate all generated feature maps. Subsequently, we apply a 1D max-pooling and $L_2$ normalization to the feature maps to generate word embeddings, $\textbf{W} \in \mathbb{R}^{256 \times (T-1)}$. Moreover, we incorporate an additional 1D max-pooling to compute the final caption embedding, $\textbf{c} \in \mathbb{R}^{256}$. In addition, another projection network, as illustrated in Figure~\ref{fig:projection_head} (b), is designed to map the global image features, $\bar{\textbf{v}} \in \mathbb{R}^{512}$, into a shared embedding space, $\textbf{v} \in \mathbb{R}^{256}$.

To increase the inherent relationships embedded within the local image features, $\bar{\textbf{R}} \in \mathbb{R}^{256 \times 14 \times 14}$, we introduce an intra-modal interaction module (IMIM). The proposed IMIM, as depicted in Figure~\ref{fig:word_region_cl}, refines local features to generate more informative representations by capturing fine-grained local patterns. It also contributes in filtering out irrelevant information, thereby improving the model's robustness to variations. In this module, we first normalize~\cite{batch_norm} the input features. Afterward, self-attention (SA)~\cite{transformer} mechanism is applied, where the keys, queries, and values are learned via 1x1 convolutions. The SA layer improves intra-modality relationships effectively, by capturing long-range dependencies of $\bar{\textbf{R}}$. It is noteworthy that in our experiment we also applied the proposed IMIM module to the textual branch. However, due to the strong intra-modal interactions, that are already present within the contextual embeddings of transformer-based foundation models~\cite{align, BLIP, bert, roberta}, the performance of our model remains the same. Hence, we skip it. 

\subsection{Face caption alignment module (FCAM)}
We design this module to learn well-aligned global and local features for both the visual and semantic modalities. The training objectives of this module are as follows: 1) CICL, designed to find relationships between the face image and its associated caption, 2) WRCL, designed to learn fine-grained alignments between image sub-regions and word embeddings. 3) IMCL, for improving feature representation within each modality and 4) an identity loss.

\paragraph{Caption-image contrastive loss (CICL)}
To learn aligned global image and caption embedding, similar to the approach in~\cite{mgca, gloria, convirt}, we design a caption-image contrastive objective between the visual and textual domains. Our CICL aims to maximize the similarity score between true face-caption pairs against random pairs.  We pass the input face image, $\textbf{x}_f$, through an image encoder and then the proposed projection network to generate the global image features, $\textbf{v}$. In parallel, we compute the contextualized caption embedding, $\textbf{c}$, from the output of the text encoder and the 1D-CNN-based projection network. We encode a mini-batch of $B$ input caption-image pairs ($\textbf{x}_f$, $\textbf{x}_c$) to generate global caption-image embeddings ($\textbf{v}$, $\textbf{c}$). Next, we compute the face-to-caption contrastive loss, $L^{(f2c)}$, for the $i^{th}$ pair in the similar way to the InfoNCE loss~\cite{infoNCE}:

\begin{equation}
\small 
L^{(f2c)}_{i} = - \log \frac{ \exp(\langle \textbf{v}_i, \textbf{c}_i \rangle / \tau) }{ \sum^{B}_{k=1} \exp (\langle \textbf{v}_i, \textbf{c}_k \rangle / \tau)}.
\end{equation}

where $\langle \textbf{v}_i, \textbf{c}_i \rangle = \textbf{v}_i^\top \textbf{c}_i / \| \textbf{v}_i \| \| \textbf{c}_i \|$ represents the cosine similarity. The loss drives the TGFR model to align positive face-caption pairs while effectively preserve the mutual information. In addition, we define a similar caption-to-face contrastive loss, $L^{(c2f)}$, as:

\begin{equation}
\small 
L^{(c2f)}_{i} = - \log \frac{ \exp(\langle \textbf{c}_i, \textbf{v}_i \rangle / \tau) }{ \sum^{B}_{k=1} \exp (\langle \textbf{c}_i, \textbf{v}_k \rangle / \tau)}.
\end{equation}
Here, $\tau$ represents a learnable parameter. We calculate the final CICL as the summation of these two losses.

\paragraph{Word-region contrastive loss (WRCL)}
To align sub-regional image features, $\textbf{R} \in \mathbb{R}^{256 \times 196}$, with word embeddings, $\textbf{W} \in \mathbb{R}^{256\times (T-1)}$, we begin by mapping the $\textbf{R}$ and $\textbf{W}$ into a multimodal semantic space. Then, we calculate a dot-product similarity matrix denoted as $\textbf{S} \in \mathbb{R}^{(T-1) \times 196}$, where $(T-1)$ represents the number of all tokens except [CLS] token, and 196 corresponds to the image sub-regions. Then, the similarities of each sub-region are normalized.

\begin{equation}
\small 
\bar{s}_{i, j} = \frac{exp(s_{i, j})}{\sum_{i=0}^{T-2}\exp(s_{k,j})}, \text{where}~~\textbf{S}=\textbf{W}^T~\textbf{R}.
\end{equation}

Next, we learn an attention weight vector, $\alpha$, which assigns varying weights to different image sub-regions according to their association with a specific word in the caption. To calculate this, we compute the attention weight for the $i^{th}$ word by normalizing $\bar{s}_{i, j}$ across all image sub-regions. Finally, we compute attention-weighted sub-region features, $\textbf{r}^{attn}$, using the attention weighted vector, $\alpha$.

\begin{equation}
\small 
\textbf{r}^{att}_{i} = \sum_{j=0}^{195} \alpha_{j} \times \textbf{r}_{j} ~~\text{where}~~ \alpha_{j} = \frac{
	\exp{(\bar{s}_{i, j}~ / \tau_1})} {\sum_{k=0}^{195}\exp{(\bar{s}_{i, j}~ /\tau_1})}.
\end{equation}

Here, $\tau_1 \in \mathbb{R}$ is the temperature parameter. Now, to find the attention-guided matching score between all words and attention-weighted sub-regions features, we define the following matching function, $f$. 

\begin{equation}
\small 
f(\textbf{x}_{f}, \textbf{x}_{c}) =  \log~( \sum^{T-1}_{i=1}\exp(\langle \textbf{r}^{attn}_i, \textbf{w}_i \rangle / \tau_2)^{\tau_2}.
\end{equation}

Here, $\tau_2$ is another hyperparameter and ($\textbf{x}_{f}, \textbf{x}_{c}$) is the input face-caption pair. Similar to approaches in~\cite{fang2015captions, damsm}, we define the word-region contrastive loss as the negative log posterior probability that aims to maximize the posterior probability of the attention-weighted region features. The contrastive loss for image sub-regions being matched with their corresponding words are given below: 

\begin{equation}
\small 
L_{WRCL}^{r|w} =\sum_{i=1}^{B}-log  \frac{
	\exp{(f(\textbf{x}_{fi}, \textbf{x}_{ci})~/ \tau_3})} {\sum_{k=1}^{B}\exp{(f(\textbf{x}_{fi}, \textbf{x}_{ck})~/ \tau_3})}.
\end{equation}

In addition, we also minimize $L_{WRCL}^{w | r}$. The final loss, $L_{WRCL}$, is the summation of these two loss.

\paragraph{Intra-modal contrastive loss (IMCL)}
While CICL and WRCL effectively align caption-image and word-region embeddings, respectively, they failed to incorporate self-supervision within each modality. Therefore, we introduce an additional contrastive objective, called intra-modal contrastive learning (IMCL), that  introduces intra-modal self-supervision, leading to the semantic alignment between true pairs within each modality. The application of the IMCL, in our work, significantly improves image-text feature representations and also makes them more discriminative. For input captions, we pair two different captions belonging to a particular subject as true pairs $(\textbf{x}_{c1}, \textbf{x}_{c2})$. Likewise, for the visual input, we consider two different face images belonging to a subject, to form a true pair $(\textbf{x}_{f1}, \textbf{x}_{f2})$. However, in cases where a subject has only one view in the dataset, we generate two random views of the subject through random data augmentation. The following objective function represents $L_{IMCL}$ loss. 

\begin{equation}
\small 
L_{IMCL} = \frac{1}{2} ({L_{IMCL}^{v}(\textbf{x}_{f1}, \textbf{x}_{f2}) + L_{IMCL}^{c}(\textbf{x}_{c1}, \textbf{x}_{c2})}).
\end{equation}
Here, $L_{IMCL}^{v}$ represents the InfoNCE loss~\cite{infoNCE} between face image pairs $(\text{x}_{f1}, \text{x}_{f2})$ where similarity is calculated using the function, $s(\textbf{x}_{f1}, \textbf{x}_{f2}) = \langle f_v(g_v(\textbf{x}_{f1})),f_v(g_v(\textbf{x}_{f2})) \rangle$. Here, $f_v$ is the image encoder and $g_v$ is the projection network. Likewise,  $L_{IMCL}^{c}$ represents the InfoNCE loss between caption pairs.

\begin{figure}[t]
	\centering
	\includegraphics[width=0.42\textwidth]{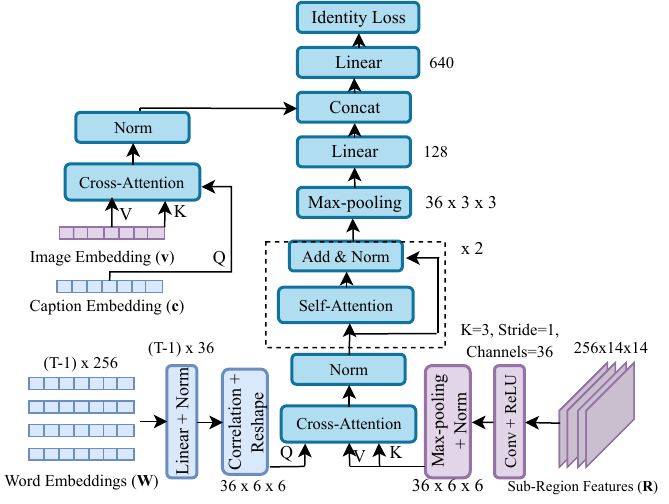}
	\caption{Block diagram of the proposed face-caption fusion module (FCFM) to capture complex fine-grained relationships and dependencies between regional features and word embeddings.}
	\label{fig:fcfm}
\end{figure}

\paragraph{Identity loss}
To increase the discriminative power of both textual and facial features, we further employ a normalized cross-entropy loss~\cite{arcface}, as our identity loss, $L_{IDL}$. This loss improves intra-class compactness while promoting inter-class dispersion. The total loss is the summation of the identity loss of both the textual and visual modalities.

\paragraph{Overall loss}
The following Eq.~\ref{equ:total_loss} delineates the objective function for the initial step of training process.
\begin{equation}
\small 
L_{total} = L_{WRCL}  + \lambda_1 L_{IDL} +  \lambda_2  L_{CICL} + \lambda_3 L_{IMCL}.
\label{equ:total_loss}
\end{equation}
Here, $\lambda_1$, $\lambda_2$, and $\lambda_3$ are the hyperparameters.

\subsection{Face caption fusion module (FCFM)}
Figure~\ref{fig:fcfm} depicts the block diagram of the proposed FCFM module. The module incorporates a cross-attention between the word embeddings, $\textbf{W}$, and regional image features, $\textbf{R}$. This cross-attention mechanism~\cite{transformer} allows the TGFR to put attention on relevant image regions while processing a specific word. A self-attention~\cite{transformer} block is applied twice to the cross-modal feature which helps the TGFR to capture inherent dependencies and interactions within the input features. Afterward, a max-pooling layer is applied to the cross-modal features before they are fed into a linear layer. To capture the coarse-grained associations between global image-caption features, we employed an additional cross-attention mechanism~\cite{transformer}, which is then followed by a normalization layer~\cite{Ba_2016}. Lastly, the output of the linear layer is concatenated with the global cross-modal features before being fed into the final linear layer.

\begin{table}[]
	\centering
	\footnotesize
	\setlength{\tabcolsep}{4pt}
	\caption{A comprehensive list of face-to-image datasets.}
	
	\begin{tabular}{cc cc}
		\toprule[1.5pt]
		Dataset                            &Public       &Images         &Annotations\\\hline
		Face2Text~\cite{face2text}         &Yes          &10,177         &1$\sim$ \\
		MM CelebA-HQ~\cite{MMCelebA} 	   &Yes          &30,000         &10  \\
		FFHQ-Text~\cite{FFHQ}              &Yes          &760            &9 \\ 
		CelebA-Dialog~\cite{celeba_dialog} &Yes          &202,599        &$\sim$5 \\
		SCU-Text2face~\cite{text2face}     &No           &1,000          &5 \\ 
		\bottomrule[1.5pt]
	\end{tabular}
	\label{table:t2f_datasets}
	\vspace{-2mm}
\end{table}

\section{Experiments}  \label{sec:expriments}
\begin{table*}[t]
	\centering
	\footnotesize
	\setlength{\tabcolsep}{5pt}
	\caption{Comparison between the proposed TGFR framework and the baselines for 1:1 verification rate and Rank-1 Identification.}
	\begin{tabular}{c cccc| cccc| cccc}  
		\toprule[1.5pt]
		\multirow{2}{*}{Methods} &\multicolumn{4}{c|}{MMCelebA~\cite{MMCelebA} Ver. (\%)} &\multicolumn{4}{c|}{Face2Text~\cite{face2text} Ver. (\%)} &\multicolumn{4}{c}{CelebA-Dialog~\cite{celeba_dialog} Ver. (\%)} \\\cline{2-5} \cline{6-9} \cline{10-13}\rule{0pt}{2ex} 
		
		&1e-5 &1e-4 &1e-3 &Id.(\%)  &1e-5 &1e-4 &1e-3 &Id.(\%) &1e-5 &1e-4 &1e-3 &Id.(\%) \\ \midrule
		
		Only AdaFace~\cite{adaface} &59.90 &70.75 &81.35 &16.43	&54.65	&57.90 &63.30  &8.38	&4.29 &10.32 &19.38 &1.01 \\
		
		BiLSTM-FLF  &66.20 &78.60 &85.0   &16.43 	&57.40  &62.40 &69.0  &8.38				 &17.47 &31.66 &48.60 &4.77  \\
		BERT-FLF    &65.50 &78.0  &85.80  &16.43 	&60.60  &63.0  &70.40 &8.38			     &22.78 &33.51 &49.31 &5.96 \\
		CGFR~\cite{cgfr}        &65.80 &81.0  &87.60  &18.90 	&61.60  &65.0  &71.80 &9.97   			 &22.36 &34.88 &50.39 &7.39 \\
		Ours &\textbf{68.20} &\textbf{81.0} &\textbf{88.20} &\textbf{21.86} 	&\textbf{64.29}&\textbf{67.81} &\textbf{74.52} &\textbf{14.33} 	&\textbf{25.50} &\textbf{37.20} &\textbf{53.0} &\textbf{10.49} \\\midrule
		
		Only ArcFace~\cite{arcface} &49.47 &55.71 &66.56 &24.65  	&46.10 &52.14 &59.09 &8.38 		 &4.90 &8.11 &14.13 &5.96 \\
		
		BiLSTM-FLF  &62.50 &71.96 &78.73  &16.43  					&61.02  &64.21 &72.0  &8.38      &14.13 &21.77 &38.65 &4.17 \\
		BERT-FLF    &63.89 &72.83 &79.89  &16.43  					&61.20  &63.96 &72.67 &8.38      &15.77 &25.23 &40.82 &12.70 \\
		CGFR~\cite{cgfr} 		&66.32 &78.05 &82.23  &28.59  		 			&63.50  &65.92 &74.48 &16.64     &16.90 &26.83 &42.45 &18.06 \\
		
		Ours &\textbf{67.72} &\textbf{78.73} &\textbf{84.47} &\textbf{33.94}  &\textbf{64.28} &\textbf{67.06}&\textbf{76.85} &\textbf{21.21}  &\textbf{19.02} &\textbf{27.84} &\textbf{44.64} &\textbf{22.0 } \\
		\bottomrule[1.5pt]
	\end{tabular}
	\label{table:eval_fr}
	\vspace{-2mm}
\end{table*}
\begin{figure*}[t]
	\centering
	\includegraphics[width=0.84\textwidth]{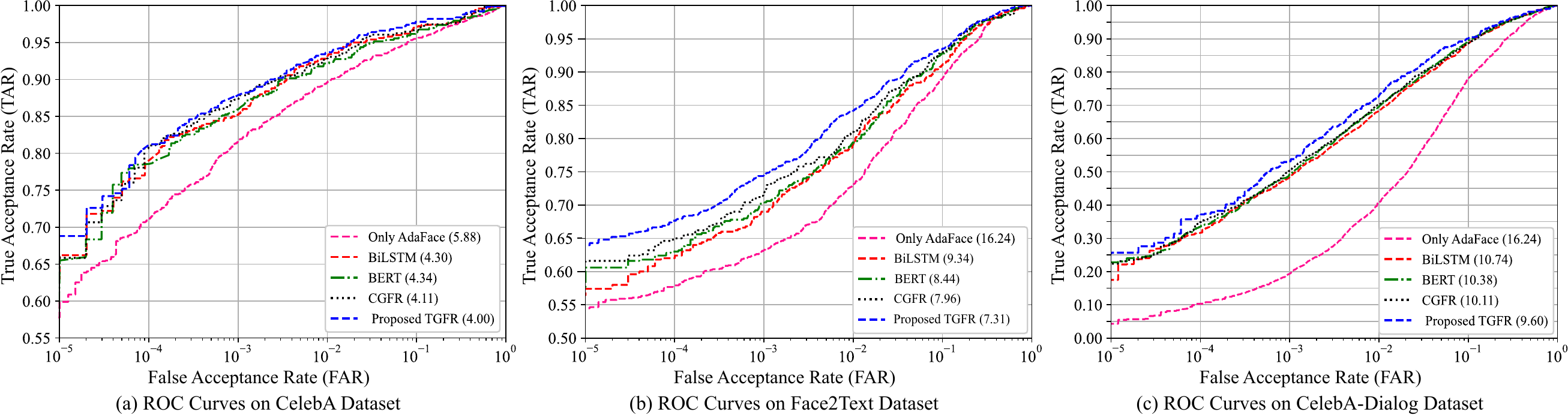}
	\caption{ROC curves depict the performance comparison of the proposed TGFR with other methods in terms of 1:1 verification rates.}
	\label{fig:adaface_eval}
	\vspace{-2mm}
\end{figure*}

\subsection{Datasets and baselines}
Table~\ref{table:t2f_datasets} presents a list of all available face-caption datasets. It is worth noting that the SCU-Text2face~\cite{text2face} dataset is not publicly available and FFHQ-Text~\cite{FFHQ} has only 760 pairs. Other face-caption datasets also have a limited number of face-caption pairs. This limitation arises due to the complexity of annotating face images, as mention in Section~\ref{sec:intro}. In this work, we conduct experiments on three publicly available face-caption datasets, namely Multi-Modal CelebA-HQ~\cite{MMCelebA}, Face2Text~\cite{face2text}, and CelebA-Dialog~\cite{celeba_dialog}. We consider the following baselines:

\textbf{BiLSTM-FLF} The bidirectional LSTM~\cite{Schuster_1997} is utilized as a text encoder and trained with an identity loss. The feature-level fusion scheme is applied using a linear layer.

\textbf{BERT-FLF} The pre-trained BERT~\cite{bert} is employed as a text encoder and fine-tuned with an identity loss. Also, a linear layer is used for the FLF scheme.

\subsection{Implementation details}
Our TGFR model has two steps training process. In the initial step, we finetune the ALIGN encoder~\cite{align} and train the FCAM module, utilizing the objective function outlined in Eq.~\ref{equ:total_loss}. We empirically set the hyperparameters $\lambda_{1} = 100$, $\lambda_{2} = 2$, and $\lambda_{3} = 1$. The ALIGN encoder~\cite{align} is finetuned for 20 epochs using an Adam optimizer\cite{Kingma_2014} with a weight decay of 0.01, and a batch size of 16. In parallel, for the projection network of both branches, we employ another Adam optimizer~\cite{Kingma_2014}, initialized with a learning rate of $0.001$. Furthermore, for the ArcFace~\cite{arcface} image encoder, the backbone iResNet18 network was pre-trained on the  MS1MV3 dataset~\cite{Deng2019b} whereas, for the AdaFace~\cite{adaface} image encoder, the same backbone was pre-trained on the WebFace4M dataset~\cite{Zhu2021webface}.  In the subsequent step, the FCFM module is trained for an additional 36 epochs. We employed an SGD optimizer with a weight decay of 1e-4 and a momentum of 0.9. The initial learning rate is 0.1 and is divided by 10 after the 6th and 24th epochs.

\subsection{Evaluation on face recognition}
In Table~\ref{table:eval_fr}, we conduct experiments to evaluate the performance of our TGFR framework for both face verification and identification. As depicted in Figure~\ref{fig:adaface_eval}, our framework substantially improves the verification rate (VR) by utilizing textual descriptions as compared to the baseline approaches and the pre-trained AdaFace model~\cite{adaface} across all three experimental datasets. For the MMCelebA dataset~\cite{MMCelebA}, our framework achieves a VR of 68.20\% at TAR@FAR=1e-5, marking a 3.52\% improvement over CGFR~\cite{cgfr} and a remarkable 12.17\% improvement over the AdaFace model. Moreover, for the identification task, our proposed TGFR attains a Rank-1 accuracy of 21.86\%, surpassing CGFR~\cite{cgfr} by 13.54\% and the baselines by 24.84\%. Similarly, on the Face2Text dataset~\cite{face2text}, our TGFR achieves a 4.18\% higher VR at TAR@FAR=1e-5 and a 30.43\% higher Rank-1 accuracy compared to CGFR~\cite{cgfr}. The CelebA-Dialog dataset~\cite{celeba_dialog} is more challenging than the other two datasets. In this dataset, the proposed TGFR remarkably improves the Rank-1 accuracy of pre-trained AdaFace from 1.01\% to 10.49\%.

We also perform experiments on the TGFR using the pre-trained ArcFace encoder~\cite{arcface} across all three datasets. In the MMCelebA dataset~\cite{MMCelebA}, TGFR achieves a VR of 67.72\% at TAR@FAR=1e-5, achieving a 2.07\% improvement over CGFR and a 26.95\% improvement over the pre-trained ArcFace model. In addition, for the identification task, TGFR achieves a Rank-1 accuracy of 33.94\%, outperforming CGFR by 15.76\% and the baseline models by 51.59\%. Similarly, on the Face2Text dataset~\cite{face2text} and the CelebA-Dialog dataset~\cite{celeba_dialog}, the proposed TGFR, using ArcFace encoder, demonstrates superior performance compared to other approaches. These results validate the generalizability and robustness of our TGFR framework.

\begin{table}[t]
	\centering
	\footnotesize
	\setlength{\tabcolsep}{4pt}
	\caption{Ablation experiments on the objective function of the proposed FCAM module using the MMCelebA dataset.}
	
	\begin{tabular}{cccc ccc}
		\toprule[1pt]
		Identity &CICL &WRCL &IMCL  &1e-5 &1e-4 &1e-5 \\\midrule
		
		\checkmark &- &- &-  &64.67 &78 &86.35 \\
		\checkmark &\checkmark &- &- &65.84 &80.50 &87.17 \\
		\checkmark &- &\checkmark &- &66.64 &81.0 &87.70 \\ 
		\checkmark &- &- &\checkmark &65.23 &78.60 &87.00 \\
		\checkmark &\checkmark &\checkmark &\checkmark  &\textbf{68.20} &\textbf{81.0} &\textbf{88.20} \\
		\bottomrule[1pt]
	\end{tabular}
	\label{table:abs_fcam}
\end{table}

\subsection{Ablation study}
\paragraph{Analysis of FCAM}
We analyze the role of each loss function in the proposed FCAM module, as shown in Table~\ref{table:abs_fcam}. In all the experiments conducted on the MMCelebA dataset~\cite{MMCelebA} using the AdaFace~\cite{adaface} encoder, the identity loss is included, which is pivotal for generating discriminative features. From Table~\ref{table:abs_fcam}, it is observed that the FCAM module without any contrastive loss experiences a decrease in VR. This proves the necessity for both intra- and inter-modal contrastive losses. Furthermore, under the evaluation metric of TAR@FAR=1e-5, WRCL demonstrates an improvement from 64.67\% to 66.64\% over the performance of the identity loss alone. In contrast to the 2.96\% improvement achieved by WRCL, we also note a 1.78\% improvement achieved by CICL and a 0.86\% improvement achieved by IMCL. These results signify the idea that each contrastive loss has a positive impact on VR performance. By combining all four components, we observe the highest VR, surpassing the identity loss alone by 5.16\%.

\begin{figure}[t]
	\centering
	\includegraphics[width=0.36\textwidth]{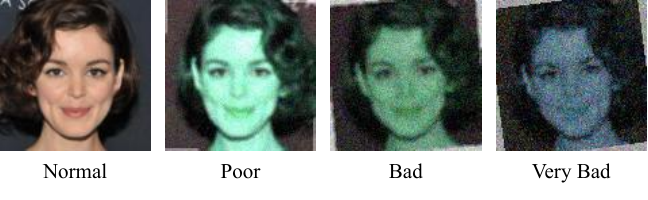}
	\caption{Face images of a subject across different image qualities.}
	\label{fig:effect_on_quality}
	\vspace{-2mm}
\end{figure}

\paragraph{Analysis of text encoder}
We investigate the impact of different foundation models, such as CLIP~\cite{clip}, and ALIGN~\cite{align} on the proposed TGFR model in Table~\ref{table:abs_text_encoder}. While the VR of all the CLIP-like foundation models is similar, they outperforms the BERT~\cite{bert}. Therefore, in TGFR, we adopt the ALIGN model~\cite{align} as our text encoder.

\begin{table}[t]
	\centering
	\footnotesize
	\setlength{\tabcolsep}{5pt}
	\caption{Ablation experiments on the text encoders}
	\begin{tabular}{c ccc}
		\toprule[1pt]
		Encoders &1e-5 &1e-4 &1e-3 \\\midrule 
		BERT-base~\cite{bert} &66.0 &81.0   &87.60\\
		ALIGN~\cite{align}    &68.20 &81.0  &88.20\\
		CLIP~\cite{clip}   	  &67.65  &81.0 &87.96 \\ 
		BLIP~\cite{BLIP}      &68.20 &81.0  &88.20  \\
		GroupViT~\cite{groupvit}  &68.0 &81.0  &88.06 \\
		FLAVA~\cite{flava} &67.90 &81.0  &88.0 \\
		\bottomrule[1pt]
	\end{tabular}
	\label{table:abs_text_encoder}
	\vspace{-4mm}
\end{table}

\begin{table}[t]
	\centering
	\setlength{\tabcolsep}{4pt}
	\caption{Comparison of 1:1 VR on TGFR framework across various image qualities.}
	\footnotesize
	\begin{tabular}{ccc cccc}
		\toprule[1pt]
		Score &Level &Method &EER &1e-5 &1e-4 &1e-3 \\\midrule 
		\multirow{3}{*}{4} &\multirow{3}{*}{normal} 	&AdaFace~\cite{adaface} &5.88 &59.90 &70.75 &81.35 \\
		& 												&CGFRR~\cite{cgfr}  &4.11 &65.80 &81.0  &87.60\\ 
		& 												&TGFR    			&4.00 &68.20 &81.0  &88.20 \\\midrule 
		
		\multirow{3}{*}{3} &\multirow{3}{*}{poor}    	&AdaFace~\cite{adaface} &10.59 &35.66 &41.41 &53.41 \\
		& 											 	&CGFRR~\cite{cgfr}  &9.88  &41.50 &44.14 &61.08\\ 
		& 											 	&TGFR    			&9.11	&44.80 &46.61 &63.12\\\midrule 
		
		\multirow{3}{*}{2} &\multirow{3}{*}{bad}   		&AdaFace~\cite{adaface} &16.30 &23.75 &31.55 &40.10 \\
		& 												&CGFRR~\cite{cgfr}     &15.10 &31.23 &35.85 &48.35  \\ 
		& 												&TGFR    				&14.68 &32.31 &37.20 &50.92 \\\midrule 
		
		\multirow{3}{*}{1} &\multirow{3}{*}{very bad}   &AdaFace~\cite{adaface} &19.06 &11.67  &18.98 &27.77  \\
		& 												&CGFRR~\cite{cgfr}     &18.78 &21.39  &29.95 &38.39  \\ 
		& 												&TGFR    				&17.65 &24.40  &31.0  &39.60  \\
		
		\bottomrule[1pt]
		\vspace{-2mm}
	\end{tabular}
	\label{table:abs_image_quality}
\end{table}

\paragraph{Effect of image quality}
We investigate the impact of image qualities on the proposed TGFR model, utilizing the AdaFace model~\cite{adaface} as the image encoder, which is trained on the ``normal'' images of the MMCelebA dataset~\cite{MMCelebA}. Figure~\ref{fig:effect_on_quality} depicts a face image of a subject across various image qualities. In this study, we report the 1:1 VR using two evaluation metrics: EER, and TAR@FAR. Our analysis, presented in Table~\ref{table:abs_image_quality}, leads to several key observations. First, we observe a drastic drop in the performance of the AdaFace model as image quality degrades. For instance, when transitioning from an image quality of ``normal" (score 4) to ``very bad" (score 1), the VR of the AdaFace substantially degraded from 59.90\% to 11.67\% at TAR@FAR=1e-5.

Secondly, across all levels of image qualities, our TGFR consistently achieves significantly higher VR compared to both CGFR~\cite{cgfr} and the pre-trained AdaFace model. For instance, in the ``very bad'' image quality, the proposed TGFR achieves an EER score of 17.16\%, which is 6.40\% lower than CGFR~\cite{cgfr} and 7.99\% lower than the AdaFace model. Additionally, we get a 12.34\% improvement compared to CGFR at TAR@FAR=1e-5, and a 3.39\% improvement at TAR@FAR=1e-4. Thirdly, as the quality of the input image degrades, the performance gain in the proposed TGFR becomes higher. For example, in the case of ``very bad" quality images, the improvement in VR of our TGFR reaches 52.17\% at TAR@FAR=1e-5 which is higher than the 20.40\% improvement observed in ``poor" quality images. These results show the efficacy of textual descriptions in boosting the performance of the FR model, specifically in low-quality face images that are corrupted with noise.

\section{Conclusion}\label{sec:conclusion}
This paper presents TGFR, a framework designed to enhance the relative performance of existing FR algorithms using textual descriptions. In this framework, we present a face-caption alignment module that effectively handles the inherent heterogeneity between the visual and the semantic domain by increasing the mutual information between the local and the global features of the face-caption pair. Our FCAM employs multiple contrastive losses across different granularities to implement inter-modal alignment between cross-modal features and intra-modal alignment within each modality. Aligning image features with textual features not only ensures the capture of complete shared semantics but also helps to focus on the distinctive content of the facial image, leading to improved performance in FR tasks.

\section{Acknowledgment}
This material is based upon a work supported by the Center for Identification Technology Research and the National Science Foundation under Grant 1650474.

{\small
\bibliographystyle{ieee_fullname}
\bibliography{wacv}
}

\end{document}